\documentclass{article}
\usepackage{arxiv}

\usepackage[utf8]{inputenc} 
\usepackage[T1]{fontenc}    
\usepackage{hyperref}       
\usepackage{url}            
\usepackage{booktabs}       
\usepackage{amsfonts}       
\usepackage{nicefrac}       
\usepackage{microtype}      
\usepackage{lipsum}		
\usepackage{graphicx}
\usepackage{natbib}
\usepackage{doi}
\usepackage{multicol}
\usepackage{subfig}
\usepackage{float}
\usepackage{url}
\usepackage{graphicx,epstopdf}
\usepackage{amsmath}
\usepackage{nth}
\usepackage{xcolor}
\usepackage{subfig}
\usepackage{algorithm}
\usepackage{algpseudocode}
\usepackage{caption}
\usepackage{soul}

\usepackage{breakurl}
\usepackage{afterpage}

\title{Real-time Fuel Leakage Detection via Online Change Point Detection}

\author{ \hspace{1mm}{Ruimin Chu} \\
	School of computing technologies\\
	RMIT university\\
	Melbourne, Australia \\
	\texttt{s3912230@student.rmit.edu.au} \\
	\And
	\hspace{1mm}{Li Chik} \\
	Titan Cloud Software\\
	Carrum Downs, Australia \\
	\texttt{li.chik@titancloud.com} \\
	\AND
	\hspace{1mm}{Yiliao Song} \\
	School of Computer and Mathematical Sciences\\
	University of Adelaide\\
	Adelaide, Australia \\
	\texttt{lia.song@adelaide.edu.au} \\
	\And
	\hspace{1mm}{Jeffrey Chan} \\
	School of computing technologies\\
	RMIT university\\
	Melbourne, Australia \\
	\texttt{jeffrey.chan@rmit.edu.au} \\
	\And
	\hspace{1mm}{Xiaodong Li} \\
	School of computing technologies\\
	RMIT university\\
	Melbourne, Australia \\
	\texttt{xiaodong.li@rmit.edu.au} \\
}

\date{}

\renewcommand{\headeright}{}
\renewcommand{\undertitle}{}

\begin{document}
\maketitle

\begin{abstract}
Early detection of fuel leakage at service stations with underground petroleum storage systems is a crucial task to prevent catastrophic hazards. Current data-driven fuel leakage detection methods employ offline statistical inventory reconciliation, leading to significant detection delays. Consequently, this can result in substantial financial loss and environmental impact on the surrounding community. In this paper, we propose a novel framework called Memory-based Online Change Point Detection (MOCPD) which operates in near real-time, enabling early detection of fuel leakage. MOCPD maintains a collection of representative historical data within a size-constrained \textit{memory}, along with an adaptively computed \textit{threshold}. Leaks are detected when the dissimilarity between the latest data and historical memory exceeds the current threshold. An \textit{update phase} is incorporated in MOCPD to ensure diversity among historical samples in the memory. With this design, MOCPD is more robust and achieves a better recall rate while maintaining a reasonable precision score. We have conducted a variety of experiments comparing MOCPD to commonly used online change point detection (CPD) baselines on real-world fuel variance data with induced leakages, actual fuel leakage data and benchmark CPD datasets. Overall, MOCPD consistently outperforms the baseline methods in terms of detection accuracy, demonstrating its applicability to fuel leakage detection and CPD problems.
\end{abstract}


\section{Introduction}
Despite the increasing popularity of electric vehicles, petroleum products will still remain the primary energy source for transportation worldwide for many years to come. Incidents involving leaks or spills of petroleum from Underground Storage Tanks (USTs) can contaminate the surrounding soil and groundwater. Moreover, the release of toxic chemicals in such situations can have severe health, economic and environmental consequences for nearby residents. There are various factors that can lead to leakage, including corrosion, defective piping, improper installation and more\footnote[1]{https://www.epa.gov/sciencematters/epa-develops-national-picture-underground-storage-tank-facilities-and-leaking}. In the USA, as of Sep 2022, there have been over 568,000 confirmed cases of release of petroleum and hazardous substances from USTs, with 4,568 confirmed releases between Oct 2021 and Sep 2022 \citep{USEPA22rep}. The annual expenditure for clean-up in the USA is approximately USD 1 billion\footnote[2]{https://www.epa.gov/ust/releases-underground-storage-tanks}. 

\begin{figure*}[t]
\centerline{\includegraphics[width=0.8\textwidth]{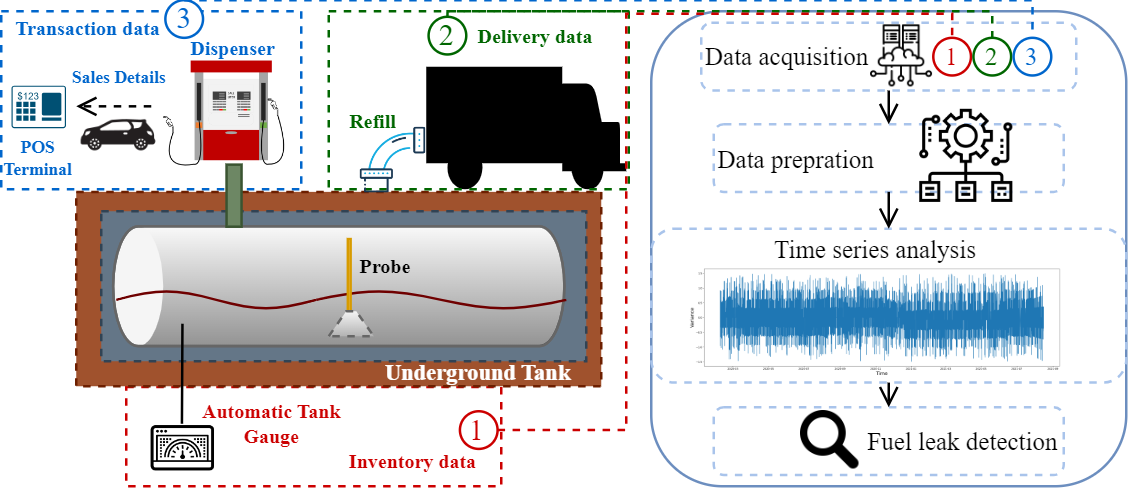}}
\caption{Layout of underground petroleum storage systems, a snapshot of activities and general fuel leakage detection process. Data from (1) inventory (2) restocking (3) customer transactions are processed for FL detection.}
\label{fig:layout}
\vspace{-4mm}
\end{figure*}

Previous data-driven studies \citep{gorawski2017tube, alayon2020time, sigut2014applying} have introduced autonomous solutions for Fuel Leakage (FL) detection to support decision-making. These approaches mainly rely on statistical inventory reconciliation which involves statistical analysis of inventory discrepancies using daily records collected over weeks \citep{USEPAGG}. They monitor the key factor, \textbf{fuel variance}, which is the discrepancy between the actual tank inventory volume and the expected fluid volume. Figure \ref{fig:layout} depicts the layout of UST systems and a commonly used FL detection framework. Data from (1) inventory (2) restocking (3) customer transactions are collected and processed to obtain fuel variance data and statistical analysis is performed to identify consistent deviations in fuel variance for FL detection.  

However, these methods typically conduct static, offline experiments that require weeks of daily data. This means that they are not able to provide a detection result until this amount of data is collected, leading to significant detection delays. A long delay in FL detection could cause severe contamination of sites and pose a threat to drinking water resources. In extreme cases, the cleanup process can span several years and incur costs amounting to millions of dollars. To ensure the safety of petrol storage and protect the surrounding environment, there is an urgent need to detect FL, more specifically those leakages occurring in storage tanks, as early as possible. Current FL detection methods are limited by their reliance on low-granularity data; whereas the increasing installation of Automatic Tank Gauges (ATGs) at service sites offers access to high-granularity data that could enable early detection. Additionally, while tanks may exhibit slightly different profiles, existing studies apply the same developed method uniformly to all tanks without considering specific tank profile information, which may impact detection accuracy. Finally, it is worth noting that fuel variance can also arise from sources other than FL, such as meter errors and evaporation losses, bringing noises to the data. These factors collectively pose challenges for accurate and timely FL detection.

In this work, we address the issue of FL detection delays by proposing a novel \emph{\textbf{M}emory-based \textbf{O}nline \textbf{C}hange \textbf{P}oint \textbf{D}etection} (MOCPD) framework that continuously processes incoming fuel inventory data and detects state changes (i.e. either from a normal state to a FL state or vice versa) in real-time. When performing detection for each individual tank instance, MOCPD not only retains a memory of historical representatives but also continuously updates this memory with the latest observations. This adaptability allows MOCPD to keep up with recent trends specific to each tank and deliver accurate and timely detections. The memory is limited by a maximum size to restrict storage costs. Additionally, MOCPD can be used with both statistical techniques and Machine Learning (ML) methods for dissimilarity measurement, making it readily deployable to engineering problems that primarily rely on traditional statistical techniques.

A key feature of MOCPD is its update phase, designed to determine \textit{when and how to update} the memory. The update phase is triggered when the buffer is filled with new data (\textit{when} to update). We evaluate three memory update schemes to determine when a new sample from the buffer should replace an existing sample in memory (\textit{how} to update). This memory update process also enables an adaptive threshold mechanism where the threshold is dynamically computed based on the samples stored in the memory. Changes are detected if the dissimilarity between the latest data and historical memory is larger than the current threshold. A memory update scheme that achieves a high level of diversity of samples can establish a good balance between the historical and recent data, enhancing the robustness of detection and achieving better recall and precision scores. 

We evaluate the performance of MOCPD on real-world fuel variance data with simulated leakages, actual fuel leakage cases and benchmark CPD datasets. The experiments demonstrate that MOCPD outperforms other baseline online CPD methods, especially for FL detection where the dataset contains many sequence instances with slightly varying profiles, more subtle state transitions and prolonged states compared to other benchmark CPD datasets. The main contributions of this paper are as follows:
\begin{itemize}
    \item We present the first study to address the fuel leakage early detection problem in a real-time setting, continuously monitoring data for each individual tank. In contrast, existing FL detection studies are performed offline and fail to account for variations in tank profiles and changes within individual tanks over time.
    \item We propose a novel memory-based online CPD framework \textbf{MOCPD}, which can operate continuously without human intervention and enable real-time detection of state changes. 
    \item We design an update phase in MOCPD to determine when and how to update the memory. With an adaptive threshold mechanism, MOCPD can achieve more accurate detection compared to baseline methods.
    \item We empirically demonstrate the effectiveness of MOCPD by comparing it with commonly used online CPD baselines on our real-world fuel variance data with induced leakage. We also carry out two case studies on real-world FL data. Finally, we demonstrate that our approach can generalise to other types of CPD problems by using several widely-used benchmark datasets.
\end{itemize}

\section{Background and Related Work}
\subsection{Fuel Leakage Detection}
\label{sec:LR_FL}
According to \citet{USEPAGG}, existing FL detection methods can be categorised into four groups: 
\begin{itemize}
\item \textbf{ATG systems}: is a volumetric-based release detection method. It analyses the physical properties of the USTs to generate an electronic signal that can be converted into the tank volume value. The processor within the console then compares calculated volumes at different times to estimate the leak rate. 
\item \textbf{Inventory reconciliation}: is another volumetric-based method that relies on inventory records to conduct a statistical analysis of inventory discrepancies. For continuous methods, it can even differentiate between line and tank leaks and can compensate for additional factors contributing to the discrepancies, such as temperature variations. However, it has strict data requirements and cannot be applied when calibrated meters and tanks or temperature-compensated data are missing.
\item \textbf{Pipeline release detection}: can be volumetric or non-volumetric. It utilises fluid flow instrumentation to observe the fluid flow rate or the static pressure in the underground piping at multiple points.
\item \textbf{Tank tightness testing}: generally a non-volumetric based method which monitors surrounding resources, such as sampling the soil or gas surrounding a UST for gasoline components, to detect leaks in the UST systems.
\end{itemize}

In this paper, we focus on the inventory reconciliation-based method using high-granularity data provided by ATGs. Inventory reconciliation methods typically track changes in fuel stock volume, referred to as \textbf{fuel variance}, over time. The fuel variance at an interval, $\text{Var}$, is calculated as:
\begin{equation}
\begin{aligned}
\text{Var} & = \text{Vol}_{\text{close}} - \text{Vol}_{\text{theoretical}}\\
 &= \text{Vol}_{\text{close}} - (\text{Vol}_{\text{open}} - \text{Vol}_{\text{sales}} + \text{Vol}_{\text{delivery}}), 
\end{aligned}
\label{eq-var}
\end{equation}
where $\text{Var}$ represents the difference between the measured closing volume ($\text{Vol}_{\text{close}}$) and theoretical inventory volume ($\text{Vol}_{\text{theoretical}}$). $\text{Vol}_{\text{theoretical}}$ can be computed using the opening volume $\text{Vol}_{\text{open}}$, sales volume over the interval $\text{Vol}_{\text{sales}}$ and delivery volume $\text{Vol}_{\text{delivery}}$.

There have been a few studies on the inventory reconciliation-based approach for FL detection. The TUBE algorithm \citep{gorawski2017tube} is an autonomous system based on the inventory reconciliation method, which incorporates data mining techniques for trend detection and data filtering. A computerised system is developed in \citet{musthafa2017automatic} with a focus on workflow, data structure and visualisation while minimising human intervention. In \citet{li2011sir}, statistical analysis is carried out over a long period and the estimated leak rate, probability of false alarm and likelihood of detection are reported. However, the method requires input measurements such as gas pressure which is collected by low-range differential pressure devices that are not available at all service stations. There are also some inventory reconciliation-based works \citep{sigut2014applying, alayon2020time, toledo2024towards} apply classical ML algorithms to the problem. These studies extract hand-crafted features from daily inventory records to represent operating days and supervised classifiers are trained on time windows of features to perform two-class classification, that classify the input data as either ''days with leaks'' or ''days without leaks''.

However, the above works generally rely on static, aggregated daily records collected over weeks or months for their experiments, which results in delayed decision-making. In contrast, our work targets FL detection in an online setting and applies online CPD methods to perform detection based on data recorded at 30-minute intervals. This enables early detection by continuously monitoring for state changes, offering a more responsive and timely solution compared to the offline methods and experimental settings of prior studies. 

\subsection{Online Change Point Detection}
\label{sec:ocpd}
CPD algorithms can be categorised into offline or online methods based on their deployment \citep{downey2008novel}. Offline algorithms analyse the entire sequence at once and identify Change Points (CPs) retrospectively, while online algorithms process data points in real-time to detect CPs as quickly as possible \citep{aminikhanghahi2017survey}. In practice, most online CPD algorithms rely on posterior data to ensure accurate detections, which creates a response delay. The requirement for 'real-time' varies depending on the specific context, ranging from milliseconds for stock trading to days for degradation pattern forecasting \citep{namoano2019online}. In the following paragraph, we discuss some state-of-the-art work in online CPD. 

\textit{Statistical}: Statistical methods can be classified into parametric and non-parametric groups, with the former making preliminary assumptions about the data and the latter not making any assumptions \citep{namoano2019online}. Bayesian Online Change-point Detection (BOCD) \citep{adams2007bayesian} is a widely used probabilistic method, which estimates the likelihood distribution of the current run length for detecting changes. The computational cost of the method is linear in the number of data points observed so far, which means it grows as the data accumulates. To address this, some extensions have been proposed, such as \citet{ranganathan2010pliss}, to restrict the runtime cost to O(n). M-Statistic \citep{li2015m} is a method based on an extension of maximum mean discrepancy (MMD) statistics. The method samples blocks of historical data, computes the quadratic-time $\text{MMD}_{u}$ of each reference block with the current block and then averages the scores. Its normalised value is compared to a predefined threshold for CPD. SEP \citep{aminikhanghahi2018real} is a real-time density ratio-based CPD algorithm that calculates divergence measures based on SEParation distance for high-dimensional time series. The method is proposed for human behaviour detection. Although density ratio-based CPD algorithms have been shown to be competitive for complex real-time problems, they often incur a large computational cost. An information gain-based method is proposed in \citet{zameni2020unsupervised} which does not require prior knowledge about the distributional of the time series and is independent of user-defined thresholds. The method specifically focuses on handling high-dimensional time series. NEWMA \citep{keriven2020newma} is a lightweight and fast detection method that uses Exponential Weighted Moving Average (EWMA) statistics with different forgetting factors to capture recent changes. QuantTree-ESWA \citep{frittoli2022nonparametric} is a non-parametric algorithm which constructs a QuantTree histogram from training samples to monitor multivariate data streams for change detection. However, the method is formulated to determine whether a sequence contains a CP and thus is not able to handle the case with multiple CPs.  

\textit{Machine Learning}: A robust Principal Component Analysis (PCA) framework \citep{xiao2019online} embeds hypothesis testing. This method computes Robust PCA in an online fashion and identifies the CPs based on the low-rank subspace and sparse error. The framework is evaluated for background subtraction in surveillance videos. A Deep Neural Network (DNN)-based algorithm \citep{gupta2022real} focuses on real-time CPD in multivariate data, particularly real-time data preprocessing. The framework has three phases: deep adaptive input normalisation \citep{passalis2019deep} to normalise data, Singular Spectrum Analysis (SSA) to remove outliers in real-time, and finally an autoencoder for CPD. An adaptive Long Short-Term Memory (LSTM)-Autoencoder is introduced in \citet{atashgahi2022memory} for unsupervised online CPD. It is a memory-free method that continuously adapts to the incoming samples and has shown great performance for multi-dimensional time series data.

\textit{Others}: LIFEWATCH \citep{faber2022lifewatch} is a lifelong learning method that applies Wasserstein distance to gauge the similarity between the probability distributions of consecutive segments. It leverages memory to keep track of previously observed distributions, allowing for the detection of CPs for recurring tasks. However, there is little implementation of memory management as it simply adds new data points to its belonging distribution until it reaches capacity. In \citet{moshtaghi2016online}, an online clustering algorithm is introduced, which can be applied to online CPD problems with adjustment. It exploits the correlation among data points to cluster them while keeping a set of decision boundaries to identify anomalous data. It comprises two main components: a distribution pool which tracks the representation of existing clusters, and the state tracker to identify when to create a new cluster. A CP can be determined when there is a need to create a new cluster.

When studying existing online CPD methods, we have noticed several limitations that require consideration when approaching the FL detection problem:
\begin{itemize}
  \item Some methods, e.g. M-statistic \citep{li2015m}, though utilising normalised dissimilarity scores, generate inconsistent score ranges for different sequences. It is impractical to choose different threshold values for various tank samples and an adaptive threshold system would be a better option for our problem. 
  \item While lightweight methods do not rely on memory (e.g. online clustering \citep{moshtaghi2016online}, SEP \citep{aminikhanghahi2018real}), they tend to have higher false alarms rates, caused by their sensitivity to single outliers.
  \item Some methods employ a memory system to store historical data, allowing them to reference that data to determine if the current distribution has changed. However, existing methods lack proper memory management that they either require storing the entire history of the current distribution (e.g. \citet{adams2007bayesian, li2015m}) or they set a memory size limit and add data until it is full, without any future replacement (e.g. \citet{gupta2022real, faber2022lifewatch}).
\end{itemize}

Based on the above findings, we decide to incorporate a size-constrained memory that will be periodically updated in the proposed MOCPD which also enables an adaptive threshold system. In this way, MOCPD can reduce false alarm rates while achieving memory efficiency.   
 
\section{MOCPD}
\subsection{Problem Formulation}
\label{sec:formulation}
We consider the streaming data from each tank as one sequence sample and each sequence is denoted as $X = \left \{x_{1}..., x_{i-1}, x_{i}, x_{i+1}, ... \right \}$, where $x_{i}$ represents fuel variance (introduced in Section \ref{sec:LR_FL}) within the $i^{th}$ 30-minute timeframe. Online CPD algorithms analyse fuel variance streams in real time and conduct CP inference while processing the data. For MOCPD, it takes the latest window $W_{i} = \left \{x_{i}, ..., x_{i+w-1} \right \}$ where $w$ is the window size, and determines if a state change has occurred for time $i$. If the dissimilarity score $s_{i}$, which measures the difference between the latest window and the current distribution, is greater than the threshold $T$, a CP is detected at time $i$. MOCPD aims to approximate the set of CPs reflecting state changes either from normal to FL states or vice versa and operates continuously without human intervention. The algorithm has no prior knowledge of the number of CPs.

\subsection{MOCPD Overview}
\begin{figure}[h!]
  \centering
    \includegraphics[width=0.7\columnwidth]{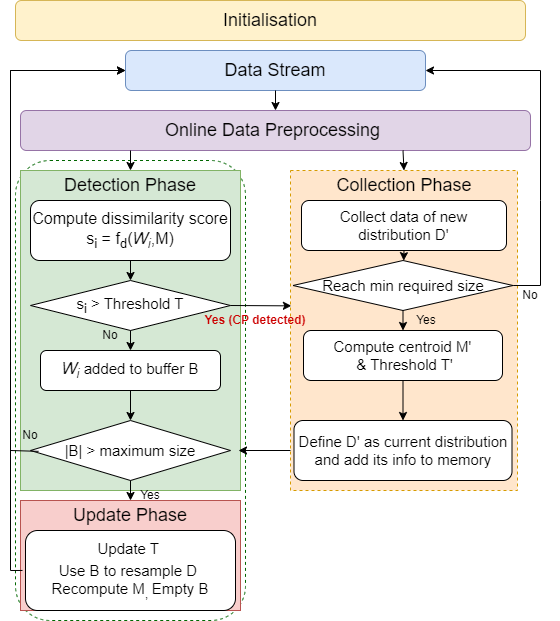}
  \caption{Pipeline of MOCPD.}
  \label{fig:workflowedge}
\end{figure}

Figure \ref{fig:workflowedge} presents the workflow of MOCPD, which comprises the following three stages: 
\begin{enumerate}
    \item \textit{Detection Phase:} measures dissimilarity between the latest window and the centroid of the current distribution. These dissimilarity measures are then compared against the threshold to determine if a CP is detected.
    \item \textit{Update Phase:} is conducted periodically to update the samples of the current distribution that are stored in the memory and the threshold. It is activated only when the buffer is filled with recent data.
    \item \textit{Collection Phase:} is initiated only when a CP is detected to gather sufficient sample data that represents the new distribution.
\end{enumerate}
Compared to existing online CPD methods, MOCPD is able to: 
\begin{itemize}
    \item periodically update the memory, adjusting samples that represent the current distribution.  
    \item adaptively update the threshold for CPD thereby enhancing precision.
    \item operate autonomously without the need for human intervention, allowing the system to run continuously.
\end{itemize}
A pseudo-code of MOCPD is presented in Algorithm \ref{alg:cap}.

\begin{algorithm}
\caption{MOCPD}\label{alg:cap}
\textbf{Input:} Fuel variance data $X = \left \{x_{1}..., x_{i-1}, x_{i}, x_{i+1}, ... \right \}$\\
\textbf{Output:} Estimated set of state change points $\nu$ \\
\textbf{Parameters:} Window size $w$, Maximum memory sample size $m$,  Minimum memory sample size $n$, Buffer size $b$, Stride size $r$, Threshold scale $\alpha$, Quantile probability $p$\\
\textbf{Initialisation:} \\
\hspace*{\algorithmicindent}Initialise distribution: $D \gets \emptyset$ \\
\hspace*{\algorithmicindent}Initialise buffer: $B \gets \emptyset$ \\
\hspace*{\algorithmicindent}$i \gets 1$ 
\begin{algorithmic}
\While{there is a data point $i$ to proceed}
\State $W_{i} \gets \left \{x_{i}, ..., x_{i+w-1} \right \}$
\If{$|D|$ $<$ $n-1$}
    \State $D \gets D \cup W_{i}$
\ElsIf{$|D|$ $=$ $n-1$}
    \State $D \gets D \cup W_{i}$
    \State $M \gets \frac{1}{\left | D \right |}\sum_{d_{i}\in D}^{}d_{i}$
    \State $S \gets \left \{ f_{d}(d_{i}, M) | d_{i} \epsilon D \right \}$
    \State $T \gets \alpha Q(S, p)$    
\Else
\State $s_{i} \gets f_{d}(W_{i}, M)$ 
\If{$s_{i}$ $<$ $T$}
    \State $B \gets B \cup W_{i}$ 
\Else
    \State $\nu \gets \nu \cup i$ 
    \State $D \gets \emptyset$
    \State $B \gets \emptyset$
\EndIf
\If{$|B|$ $>$ $b$}
    \State $S \gets \left \{ f_{d}(d_{i}, M) | d_{i} \epsilon D \right \}$
    \State $T \gets \alpha Q(S, p)$
    \State Update memory \Comment{refer to Section \ref{sec:update}}
    \State $M \gets \frac{1}{\left | D \right |}\sum_{d_{i}\in D}^{}d_{i}$
    \State $B \gets \emptyset$ 
\EndIf
\EndIf
\State $i \gets i + r$
\EndWhile
\end{algorithmic}
\end{algorithm}

\subsection{Online Data Pre-processing}
\label{sec:datapre}
Most existing algorithms in the field of online CPD pre-process the dataset offline \citep{gupta2022real}. However, in real-world scenarios where data is not available all at once, online pre-processing becomes necessary. In the context of FL detection, factors including unregistered measurements and probe errors can affect data quality \citep{gorawski2017tube}. Therefore, data filtering is essential to minimise noises and outliers. We employ a similar strategy as introduced in \citet{gupta2022real, lu2018detecting}, which uses Singular Spectrum Analysis (SSA) for real-time local outlier removal. SSA decomposes the time series into interpretable components and then reconstructs the input based on the decomposition. The residual is calculated as the difference between the original signal and the reconstruction. Residual values exceeding the threshold are identified as outliers and we replace them with the mean value of the previous window with a size of 10 to ensure the imputation captures a sufficient amount of recent data.

\subsection{Initialisation}
MOCPD first learns the initial distribution $D$ of fuel variance in an unsupervised manner by processing the input variance stream into overlapping windows of length $w$ with a stride of $r$. After collecting the required minimum amount of windows (this size can be determined based on the characteristics of the dataset) for a distribution, the centroid $M$ of $D$ is computed as:
\begin{align*}
M = \frac{1}{\left | D \right |}\sum_{d_{i}\in D}^{}d_{i} .
\stepcounter{equation}\tag{\theequation}\label{eq:mc}
\end{align*}
The threshold $T$ of $D$ is computed using: 
\begin{align*}
& S = \left \{ f_{d}(d_{i}, M) | d_{i} \in D \right \}, \\
& T = \alpha Q(S, p), \stepcounter{equation}\tag{\theequation}\label{eq:threshold}
\end{align*}
where $S$ is the set of all the distances between instances in $D$ and the centroid $M$ and $f_{d}$ is the dissimilarity measure estimator. $Q$ is the quantile function with $p$ as a probability value between 0 and 1, and $\alpha$ controls the scale of the threshold. Finally, $D$ and its centroid $M$, threshold $T$ are stored in the memory. 

\subsection{Detection Phase}
The detection phase is the primary phase to determine if a CP has occurred. MOCPD can be used with various dissimilarity measure estimators to compare the latest window $W_{i}$ and the centroid of current distribution $M$. The selection of dissimilarity measure $f_{d}$ depends on the CP type and application problem. In this paper, we evaluate three different measures, ranging from parametric measure specialised in the mean property to align with the characteristics of the FL problem to non-parametric measures that are adaptable to various types of CPs. The computed dissimilarity score $s_{i}$ is then compared to the threshold $T$ associated with $D$, to determine if $W_{i}$ belongs to $D$. If the system identifies a CP, it will proceed to the collection phase. If not, $W_{i}$ will be added to the buffer $B$ for later use in the update phase. 

\textbf{Mean}: The arithmetic mean is used to represent each window:
\begin{equation}
 s_{i} = (\mu _{W_{i}} - \mu _{M})^{2}.   
\end{equation}
Intuitively, the squared difference between the arithmetic mean of $W_{i}$ and the arithmetic mean of centroid $M$ is used as the score.

\textbf{Maximum Mean discrepancy (MMD)}: Maximum mean discrepancy \citep{gretton2012kernel} is a non-parametric kernel-based statistic defined by the distance between mean embeddings on a \textit{reproducing kernel Hilbert space} (RKHS). The measure has been previously applied for CPD \citep{li2015m,chang2019kernel} and has the advantage of requiring fewer assumptions on the distributions. In our scenario, given $W_{i}$, $M$ and a kernel $k$, the MMD is the distance between the means of the embedding of $W_{i}$ and $M$ and is defined as:
\begin{align*}
    & s_{i} = \text{MMD}_{k}^{2}(W_{i},M) = \left \| \mu_{W_{i}} - \mu_{M} \right \|_{\mathcal{H}}^{2} \\
    & = E_{x,x'\sim W_{i}}\left [ k(x,x') \right ] -2E_{x\sim W_{i},y\sim M}\left [ k(x,y) \right ] \\
    & \quad +E_{y,y'\sim M}\left [ k(y,y') \right ], 
\end{align*}
where x and x' are independent data points within distribution $W_{i}$, and y and y' are independent data points within distribution $M$.

\textbf{Variational AutoEncoder (VAE)}: VAE \citep{kingma2013auto} is a generative model that approximates the posterior distribution using variational inference. It models the given observation $x$ with latent variables $z$. A probabilistic encoder, a recognition model, $q_{\phi }(z|x) = N(\mu _{z}(x,\phi ), \sigma ^{2}_{z}(x,\phi ))$ is learnt to approximate the true posterior distribution via a Multi-Layer Perceptron (MLP). A probabilistic decoder $p_{\theta }(x|z)$ is constructed to compute a probability distribution from $z$ with an MLP. The encoder's parameters $\phi$ and the decoder's parameters $\theta$ are learnt simultaneously through back-propagation. The loss function is written as:
\begin{equation}
L_{VAE}(\theta,\phi) = -D_{KL}(q_{\phi}(z|x)||p_{\theta }(z))+E_{z\sim q_{\phi }(z|x)}log p_{\theta }(x|z),    
\end{equation}
where the first term denotes the KL divergence between the approximate and the true posterior, to encourage regularisation. The second term is to optimise the reconstruction of $x$ through maximising the marginal likelihood of $p_{\theta }(x)$. 

During the initialisation phase, the model is trained with windows to learn the general patterns in the initial distribution. The encoder of the trained VAE will be used to obtain the latent representation of windows. We compute the mean squared error between the mean latent variable of the window $\mu _{z}(W_{i},\phi )$ and the mean latent variable of the centroid $\mu _{z}(M,\phi )$ as the dissimilarity score:
\begin{align*}
s_{i} = \left\| \mu _{z}(W_{i},\phi ) - \mu _{z}(M,\phi ) \right\|^{2}.
\end{align*}
\subsection{Update Phase}
\label{sec:update}
Once the buffer $B$ reaches its predefined maximum size, the update phase is activated. The intuition is that data may evolve over time while this change is gradual and not large enough to qualify as a CP. Thus, it is essential to update the samples of $D$ as new data arrives while maintaining the diversity of the samples. 

We explore three memory update schemes that are suitable for our problem setting to determine which samples should be retained in memory. These algorithms include random sampling, reservoir sampling \citep{vitter1985random} and prototype sampling \citep{rebuffi2017icarl}: 
\begin{enumerate}
    \item Random sampling: we follow the scheme in \citet{bang2021rainbow} where the exemplars to be kept in the memory are sampled uniformly at random from the existing samples in the memory and the incoming samples in the buffer without replacement.
    \item Reservoir sampling \citep{vitter1985random}: it conducts uniform random sampling without replacement on a data stream without requiring a priori knowledge of the stream length. Specifically, the reservoir is initialized with the first $m$ elements of the stream where $m$ is the memory sample size. Then each new element is added to the reservoir with a probability of $m/i$ where $i$ is the index of the element.
    \item Prototype Sampling \citep{rebuffi2017icarl}: selects the top $m$ samples which are closest to the mean of the distribution.
\end{enumerate}

During this update phase, the threshold is also adjusted based on the samples in memory, enabling an adaptive threshold mechanism. When investigating methods employing an adaptive threshold system, we have observed a common limitation. Since thresholds are typically updated based on the dissimilarity scores, they tend to synchronise with the similarity scores. This means that when there is an actual CP, if the threshold changes immediately in response to the score, it can quickly surpass the score and ultimately lead to the system missing the actual CPs. For MOCPD, we deliberately introduce a small delay when updating the threshold, by conducting the threshold update before resampling. This ensures that for actual CPs, the threshold will not exceed the dissimilarity score, to maintain a better recall score. Once resampling is finished, $M$ is updated accordingly. 
\subsection{Collection Phase}
\label{sec:collect}
The collection phase is activated when a CP is detected. The purpose of this phase is to gather sufficient data to reflect the newly detected distribution $D'$. Similar to the process during initialisation, it computes the corresponding centroid $M'$ and threshold $T'$. Finally, the algorithm defines $D'$ as the new current distribution so that when it returns to the detection phase, the new upcoming window will be compared against to this new current distribution. When using VAE for dissimilarity measure, the model will be retrained with samples of $D'$. 

\section{Experiments}
\subsection{Data}
The dataset used in this paper is collected from real service stations, including 30-minute interval inventory data, sales and delivery records. In total, we have collected 160 tank samples within the period of 2019-2022 from different states in Australia. The fuel variance (as introduced in Section \ref{sec:LR_FL}) is constructed using the above-mentioned three data sources. We use simulated data for experiments due to the rarity of real FL cases. Since the simulation of leakage is added to the real data sequences, the dataset retains the noise and unexpected patterns that can occur in the real scenarios. This allows us to provide insights into how the noisy data can affect fuel leakage detection. To simulate the tank leakage data for testing, we follow the guidelines outlined in \citet{USEPASIR, USEPAGG}, with some modifications to accommodate our data sampling rate and format. According to Environmental Protection Agency's performance standard, three groups of average leak rates are simulated, which are 0.05 gallon per hour (gph), 0.1 gph and 0.2 gph. For each average leak rate group, a leak rate is generated for each sequence sample using a uniform distribution within ± 30 percent of the average leak rate (e.g. 0.14 to 0.26 gal/hr for 0.2 gph (0.2 ± 30\%)). We follow the assumption that the leakage occurs at the bottom of the tank, meaning that the leakage happens throughout the entire leakage period. Then we randomly determine a leakage start date which is at least two months after the beginning of the sequence to allow for the study of normal state data. The stop date is randomly determined and set to be at least three months after the start date. 

For each 30-minute window during the leakage period, we compute the fuel leakage volume based on the pre-determined leak rate, using the following equation:
\begin{align*}
0.5\times lr\sqrt{\frac{h_{product level}}{h_{max}}},
\end{align*}
where $lr$ is the generated leak rate, $h_{product level}$ is the current product level and $h_{max}$ is the maximum inventory height for the month. This calculated leakage volume is then added to the raw fuel variance value. 

Temperature compensation is applied to standardise the volume to 15 degrees Celsius in order to moderate the temperature effect. Finally, we divide the data into idle period, transaction period (transactions occurred within the 30-minute interval) and delivery period (restock). We only keep the idle period data for experiments as it contains less disturbance from other sources of fuel variances. 

\subsection{Baseline}
We compare MOCPD with the following commonly used online CPD algorithms mentioned in section \ref{sec:ocpd}, which are suitable for univariate problems, on our FL detection. They are: 
\begin{itemize}
    \item \textbf{BOCD} \citep{adams2007bayesian}: a probabilistic model that computes the probability of the current run length.
    \item \textbf{M-Statistic} \citep{li2015m}: a kernel-based method which utilises MMD to measure the difference between current sample and history data.
    \item \textbf{NEWMA} \citep{keriven2020newma}: a model-free method that computes exponential weighted moving average with different forgetting factors on the data stream.
    \item \textbf{SEP} \citep{aminikhanghahi2018real}: a density-ratio based method which computes divergence measure based on Separation distance.
    \item \textbf{AE} \citep{gupta2022real}: a DNN-based model which uses an autoencoder and analyses its reconstruction error for CPD.
    \item \textbf{OC} \citep{moshtaghi2016online}: an online clustering method that has been modified for CPD.
    \item \textbf{LIFEWATCH} \citep{faber2022lifewatch}: a Wasserstein distance-based approach with memory to model multiple data distributions in an unsupervised manner.
\end{itemize}
     
\subsection{Setup}
We implement MOCPD in Python 3.8.10, using Tensorflow \citep{abadi2016tensorflow} and Keras \citep{chollet2015keras}. The experiments are run in an environment of AMD EPYC 7502 CPU processor and NVIDIA A100 GPU. For all the algorithms, we perform the pre-processing step as described in section \ref{sec:datapre}. The hyperparameters (including the ones introduced in Algorithm \ref{alg:cap} and VAE's setup) are selected either based on the grid search or the data properties of our FL data. Methods with a data collection phase are set to transit back to the detection phase after 500 samples. In MOCPD, we set $r$ = 10, $p$ = 0.975, $\alpha$ = 4, $m$ = 75, memory update scheme as random sampling after hyperparameter tunning. The update phase is activated every 150 samples. The encoder and decoder are set with a single hidden layer consisting of 4 neurons using the ELU activation function. We use the RMSprop optimiser to train VAE for 100 epochs with an initial learning rate of 0.01. Hyperparameters in baseline models are set with the default settings or selected via grid search, and the best results are reported for each method.

\subsection{Evaluation Metrics}
We consider a variety of performance measures that are relevant to FL criteria and online CPD problems. If a state change is detected within the next 10 days then it is considered to be a true positive instance.
\begin{itemize}
  \item \textbf{Recall}: also referred as True Positive Rate. In our context, it represents the proportion of actual state changes that are correctly recognised. 
  \item \textbf{Precision}: another key criterion in the context of FL \citep{USEPAGG}. It is measured by dividing the number of correctly identified CPs by the total number of detections made.
  \item \textbf{F2-score}: a modified version of the F1-score, which is the weighted harmonic mean of precision and recall. It is calculated as:
\begin{equation}
\label{eq:fscore}
F_{\beta }=(1+\beta ^{2})\frac{precision\cdot recall}{(\beta ^{2}\cdot precision)+recall},
\end{equation}  
with $\beta = 2$, to place more emphasis on recall.
  \item \textbf{Detection delay}: measures the deviation of the correctly detected CP from the position of actual CP. It is computed by taking the average of the absolute difference between the timestamp when a decision is made and the actual CP timestamp. A shorter detection delay is preferred.
\end{itemize}

\section{Empirical Evaluation}
Overall, MOCPD's performance is competitive against the baseline methods in FL detection especially in terms of detection accuracy. We first evaluate experiments' results on simulated FL data with an average leak rate of 0.2 gph. This is the largest simulated leak rate in our study which produces the most obvious trend change. For the non-deterministic methods, the average scores from five runs are reported. F2-score is used as the primary measure because it slightly favours recall over precision, as missing actual leakages is worse than having more false alarms. 

As shown in Table \ref{tab:02res}, MOCPD achieves the highest F2-scores, up to 0.61. When using MEAN and MMD as the dissimilarity measures, it outperforms the other baseline algorithms. The performance of MOCPD-VAE is worse than the other two measures. This is likely due to its dependency on data quality and VAE being overly complex for the problem. Among the baseline methods, LIFEWATCH obtains the highest F2-score of 0.51. However, it is not as good as MOCPD due to its lower recall rate. BOCD and NEWMA can achieve a recall score above 0.5; however, their precisions are much lower, indicating that they are prone to noises and local trends that exist in our real-world data. Lightweight memory-free methods, such as NEWMA, SEP and OC, do not perform well in our problem setting where the state transitions are subtle, bringing challenges in differentiating them from noises. In terms of detection delay, MOCPD-based methods generally lag behind other baselines, which could be partially affected by the availability of idle period data. For example, busy sites with more transactions have less data that meets the conditions for analysis and thus lead to longer delays. However, these delays remain far shorter than the general turnaround time in practice, which is typically over 20 days for inventory reconciliation-based methods employed in industries\footnote{http://www.nwglde.org/methods/sir\_quantitative.html}. In summary, while MOCPD does not excel in every metric, it strikes a strong balance between precision and recall and effectively handles noise and outliers, making it a robust choice for real-time FL detection where detection accuracy is crucial.
\begin{table*}[h!]
\caption{Results of MOCPD with MEAN, MMD and VAE used for dissimilarity measure and the baselines with different window sizes on \textbf{0.2 gph} FL data. Recall, Precision (Prec), F2-score (F2) and Detection Delay (DD) in days are reported.} 
\setlength{\tabcolsep}{2.5pt}
\centering
\resizebox{\textwidth}{!}{%
\begin{tabular}{l|lll|l|lll|l|lll|l|lll|l}
 & \multicolumn{4}{c|}{$w$=50} & \multicolumn{4}{c|}{$w$=75} & \multicolumn{4}{c|}{$w$=100} & \multicolumn{4}{c}{$w$=125} \\
\hline
Method            & Recall & Prec & F2 & DD & Recall & Prec & F2 & DD & Recall & Prec & F2 & DD & Recall & Prec & F2 & DD \\ \hline
BOCD              & 0.5147 & 0.1147 & 0.3019 & 2.07 & 0.5265 & 0.1275 & 0.3238 & 3.0 & 0.5235 & 0.1346 & 0.3318 & 3.92 & 0.5471 & 0.1480 & 0.3554 & 4.84 \\
M-Statistic       & 0.2972 & 0.3722 & 0.3096 & 3.70 & 0.3500 & 0.3300 & 0.3459 & 4.15 & 0.3843 & 0.2881 & 0.3602 & 4.76 & 0.3626 & 0.2585 & 0.3354 & 4.94 \\
NEWMA             & 0.4029 & 0.050 & 0.1661 & \textbf{2.02} & 0.4471 & 0.0757 & 0.2256 & \textbf{2.18} & 0.5147 & 0.1123 & 0.3000 & \textbf{2.37} & 0.5294 & 0.1308 & 0.3289 & \textbf{2.62} \\
SEP               & 0.1422 & 0.2782 & 0.1575 & 3.79 & 0.3216 & 0.2135 & 0.2920 & 4.39 & 0.4804 & 0.1837 & 0.3631 & 5.99 & 0.4706 & 0.2235 & 0.3853 & 6.62 \\
AE                & 0.2235 & 0.1863 & 0.2149 & 3.03 & 0.2461 & 0.1546 & 0.2200 & 5.01 & 0.2818 & 0.2299 & 0.2696 & 6.04 & 0.3225 & 0.1968 & 0.2860 & 7.61 \\
OC                & 0.3554 & 0.3430 & 0.3529 & 2.78 & - & - & - & - & - & - & - & - & - & - & - & - \\
LIFEWATCH         & 0.4848 & 0.6046 & 0.5048 & 2.95 & 0.5183 & 0.4450 & 0.5018 & 3.84 & 0.5488 & 0.3838 & 0.5053 & 4.80 & 0.5640 & 0.3179 & 0.4883 & 5.16 \\ \hline
\textbf{MOCPD-Mean}    & 0.6398 & 0.3899 & 0.5671 & 3.48 & 0.6820 & 0.4014 & 0.5983 & 5.00 & 0.7031 & 0.3913 & 0.6064 & 6.06 & 0.6922 & 0.3815 & \textbf{0.5952} & 7.64 \\ 
\textbf{MOCPD-MMD}     & 0.6667 & 0.3992 & \textbf{0.5879} & 3.46 & 0.7104 & 0.4024 & \textbf{0.6161} & 4.96 & 0.7052 & 0.3958 & \textbf{0.6098} & 6.25 & 0.6760 & 0.3788 & 0.5843 & 7.50\\ 
\textbf{MOCPD-VAE}     & 0.5344 & 0.3858 & 0.4961 & 2.81 & 0.5177 & 0.4115 & 0.4922 & 3.99  & 0.5677 & 0.4301 & 0.5335  & 5.31  & 0.5240 & 0.4419 & 0.5051 & 6.24 \\ \hline
\end{tabular} 
}
\label{tab:02res}
\end{table*} 

\begin{figure*}[]
  \centering  
    \includegraphics[width=0.9\textwidth]{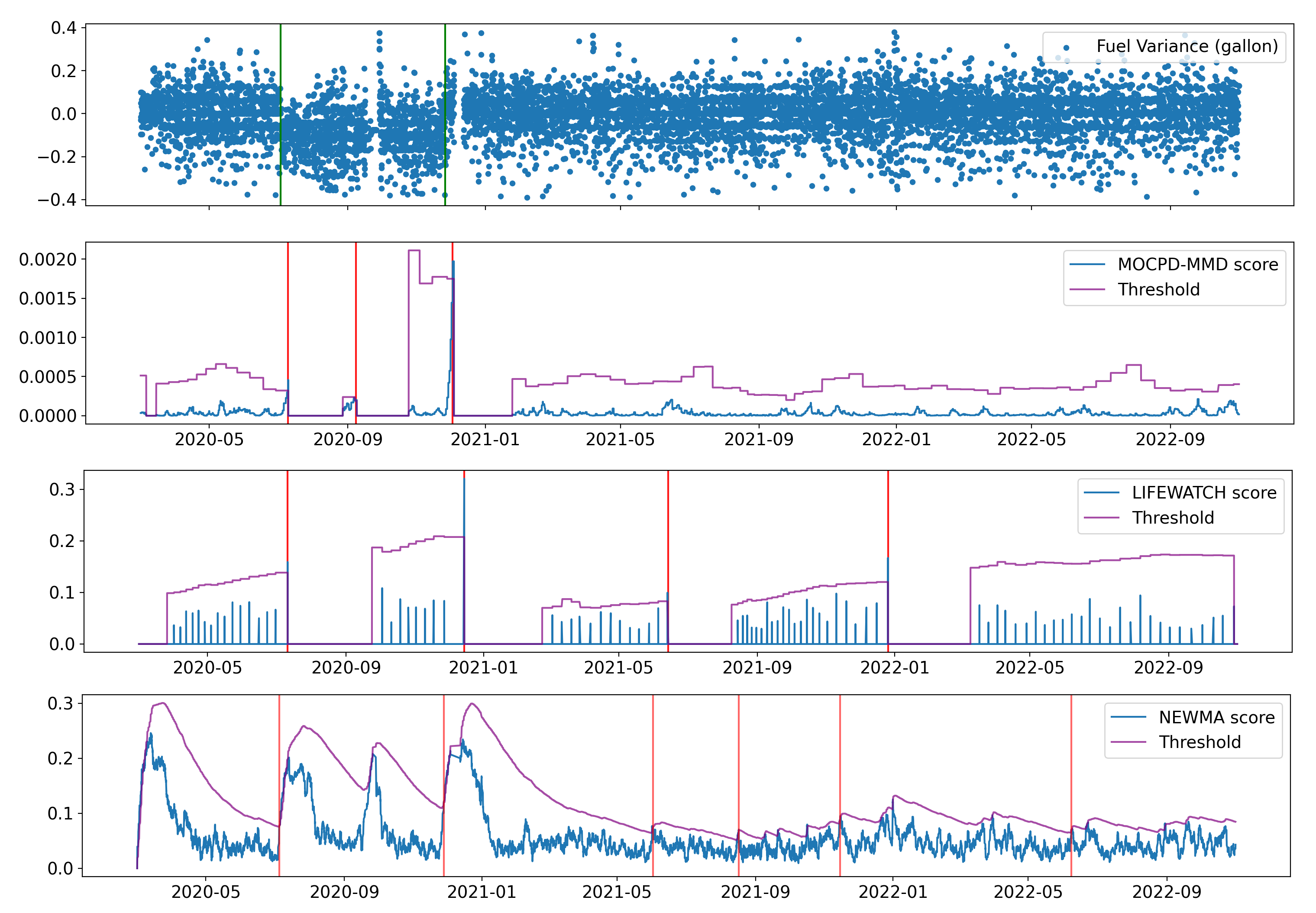}
  \caption{The dissimilarity scores and thresholds obtained by different algorithms on the 0.2gph FL sample (top plot). Green vertical lines indicate the ground truths where the leakage starts and is fixed. Red vertical lines denote the predictions made by the algorithms.}
  \label{fig:02scores}
\end{figure*}

We also plot the dissimilarity scores and threshold trends to examine how MOCPD compares with baselines. Figure \ref{fig:02scores} visualises an example of 0.2 gph fuel variance data in the top plot. The corresponding dissimilarity scores and adaptive threshold trends obtained by MOCPD-MMD and two of the baseline methods are displayed in Figure \ref{fig:02scores}. The scores for LIFEWATCH are discontinuous because the windows are non-overlapping, and a set of scores and thresholds is generated per window size. The flat trends observed in MOCPD-MMD and LIFEWATCH suggest their collection periods when there is no detection conducted. All the methods are able to capture the actual CPs. However, MOCPD-MMD generates fewer false alarms than NEWMA and LIFEWATCH. NEWMA generates the most false alarms, suggesting that it is sensitive to noises. On the other hand, MOCPD-MMD is able to respond to the CPs while being robust against outliers and noise.    

We also evaluate the runtimes of the algorithms. For each algorithm, we record the time it takes to make a decision at every step, with $w$ set to 100, i.e. the time taken to generate the dissimilarity score and determine whether it signifies a CP. The average runtimes in milliseconds for each algorithm are as follows: \textbf{NEWMA}: 0.12, \textbf{MOCPD-MEAN}: 0.21, \textbf{AE}: 0.60, \textbf{OC}: 0.70, \textbf{MOCPD-MMD}: 1.04, \textbf{BOCD}: 1.57, \textbf{LIFEWATCH}: 2.52, \textbf{MOCPD-VAE}: 2.86, \textbf{M-Statistic}: 3.62, \textbf{SEP}: 313.43. The runtimes for all methods are acceptable except for SEP, which incurs a much larger computational cost than other methods. However, given that this turnaround time is significantly smaller than the sampling rate of the data (30 minutes), this will not be a main issue in practice.

\subsection{Impact of Window Size} 
\label{sec:windowsize}
\begin{figure}[]
  \centering
    \includegraphics[width=0.8\columnwidth]{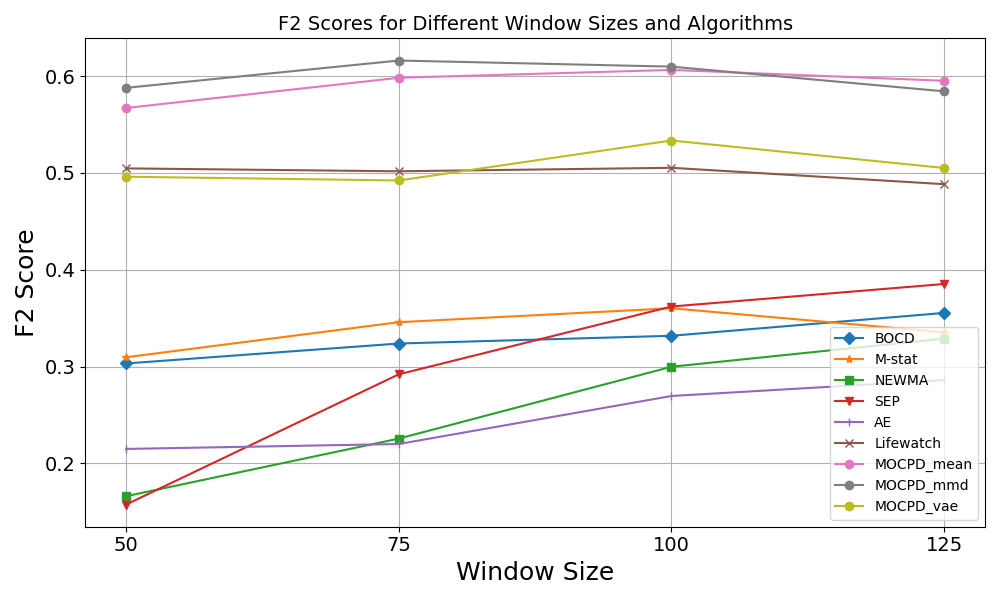}
  \caption{F2-scores for different window sizes and algorithms.}
  \label{fig:f2_ws}
\end{figure}

All the implemented methods except OC follow a window-based paradigm or can accommodate a decision delay allowance. CPD aims at identifying permanent trend changes rather than single outliers; thus, a proportion of new data is generally required to enable the algorithm to make reliable detections. Therefore, there is a trade-off between detection delay and performance and generally a larger window size will lead to a longer detection delay. We vary the window size $w$ from $\left \{50,75,100,125 \right \}$ and Figure \ref{fig:f2_ws} shows the F2-scores when algorithms use different window sizes. Most methods achieve their best performances when $w$ is above 75, with a detection delay of 2-7 days. For BOCD, NEWMA, SEP and AE, performances always improve as $w$ increases, though the speed of improvement slows down. For the remaining methods, the increment of $w$ does not necessarily guarantee a performance enhancement. In cases where F2-scores are similar, a smaller window size is preferred as it should lead to a shorter detection delay. 

\subsection{Impact of Memory Update Scheme} 
As mentioned in section \ref{sec:update}, we have investigated three memory update schemes for the update stage of MOCPD, and we compare their performances on all simulated leak rate datasets with $w = 100$ (the best-performing setting from subsection \ref{sec:windowsize}) and $m = 75$. The F2-score results are summarised in Table \ref{tab:memoryupdate}. As shown in the table, random sampling outperforms the other two memory update schemes. This is likely due to the fact that this scheme allows more frequent updates with recent data compared to the others. Among the three schemes, prototype sampling shows the worst performance as it only keeps track of those samples that are close to the centroid, limiting the diversity of the samples that the scheme retains. The model then becomes sensitive to local anomalies, thus having a much higher false alarm rate. Moreover, the scheme is computationally expensive, as the distance needs to be computed each time. Though reservoir sampling is suitable for data stream setup, the recent elements have less chance of being included in the memory, making it worse compared to random sampling. Overall, a memory update scheme that can maintain the diversity of samples for $D$ while not updating the distribution too rapidly is preferred in our problem setting. In this way, it can capture real CPs while being robust to local sudden changes. 
\begin{table*}[]
\caption{MOCPD's F2-scores on FL data with different memory update schemes.}
\centering
\begin{tabular}{c|lll|lll|lll}
\multicolumn{1}{l|}{} & \multicolumn{3}{c|}{leak rate=0.2} & \multicolumn{3}{c|}{leak rate=0.1} & \multicolumn{3}{c}{leak rate=0.05} \\
\multicolumn{1}{l|}{} & \multicolumn{1}{c}{MEAN} & \multicolumn{1}{c}{MMD} & \multicolumn{1}{c|}{VAE} & \multicolumn{1}{c}{MEAN} & \multicolumn{1}{c}{MMD} & \multicolumn{1}{c|}{VAE} & \multicolumn{1}{c}{MEAN} & \multicolumn{1}{c}{MMD} & \multicolumn{1}{c}{VAE} \\ \hline
Random               & 0.6064 & 0.6098 & 0.5335 & 0.4495 & 0.4438 & 0.3742 & 0.2057 & 0.1997 & 0.1591 \\
Reservoir            & 0.5377 & 0.5237 & 0.4391 & 0.3638 & 0.3615 & 0.2959 & 0.1253 & 0.1222 & 0.1074 \\
Prototype            & 0.3026 & 0.2888 & 0.3224 & 0.2251 & 0.2021 & 0.2461 & 0.0989 & 0.0949 & 0.1593 \\ \hline
\end{tabular} 
\label{tab:memoryupdate}
\end{table*}

\subsection{Impact of Memory Size Limit} 
To evaluate how memory size limit can affect the performance of memory-based methods, including MOCPD and LIFEWATCH, on the FL problem, we compare the results on all simulated data with $w = 100$, random sampling for the memory update scheme and vary the maximum memory sample size $m$ between $\left \{50,75,100,125 \right \}$. The F2-scores are reported in Table \ref{tab:memorysize}. We observe that the overall performance is better when $m$ is set to 75 compared to other conditions, though the differences in scores are not significant. This indicates that storing more samples does not guarantee a better detection performance; moreover, when F2-scores are similar, a smaller memory size is preferred as it is more efficient.

\begin{table*}[]
\caption{LIFEWATCH and MOCPD's F2-scores on FL data with different memory sample size limits.}
\centering
\resizebox{\textwidth}{!}{%
\begin{tabular}{c|llll|llll|llll}
\multicolumn{1}{l|}{} & \multicolumn{4}{c|}{leak rate=0.2} & \multicolumn{4}{c|}{leak rate=0.1} & \multicolumn{4}{c}{leak rate=0.05} \\
\multicolumn{1}{l|}{} & \multicolumn{1}{c}{m=50} & \multicolumn{1}{c}{75} & \multicolumn{1}{c}{100} & 125 & \multicolumn{1}{c}{50} & \multicolumn{1}{c}{75} & \multicolumn{1}{c}{100} & 125 & \multicolumn{1}{c}{50} & \multicolumn{1}{c}{75} & \multicolumn{1}{c}{100} & 125 \\ \hline
MOCPD-MEAN     & 0.5833 & 0.6064 & 0.5918 & 0.5871 & 0.4357 & 0.4495 & 0.4017 & 0.3939 & 0.2023 & 0.2057 & 0.1740 & 0.1711 \\
MOCPD-MMD      & 0.5891 & 0.6098 & 0.5882 & 0.5885 & 0.4277 & 0.4438 & 0.4005 & 0.4020 & 0.1823 & 0.1997 & 0.1715 & 0.1574 \\
MOCPD-VAE      & 0.5002 & 0.5335 & 0.4791 & 0.4517 & 0.3584 & 0.3742 & 0.3099 & 0.3096 & 0.1816 & 0.1591 & 0.1411 & 0.1309 \\ 
LIFEWATCH      & 0.5140 & 0.5053 & 0.4989 & 0.4967 & 0.3350 & 0.3560 & 0.3102 & 0.3059 & 0.1211 & 0.1866 & 0.1002 & 0.0973 \\ \hline
\end{tabular}
}
\label{tab:memorysize}
\end{table*}

\subsection{Impact of Leak Rate} 
To evaluate how MOCPD and baseline algorithms perform under FL scenarios with smaller leak rates where the trend change is less pronounced, we compare their results on 0.1 gph and 0.05 gph simulated data. The scores are summarised in Tables \ref{tab:01} and \ref{tab:005res}. As the leak rate decreases, the performances of all the algorithms deteriorate. The fuel loss caused by the low leak rate is also challenging for humans to visually identify. For 0.1 gph leakage data, MOCPD-Mean and MOCPD-MMD achieve F2-scores of 0.4495 and 0.4438, respectively, which are the best two compared to other methods. Most methods have a decrease of 0.1$\sim$0.15 in F2-scores compared to their scores for the 0.2 gph scenario. In the case of 0.05 gph data, the performance of all the algorithms becomes worse, as the trend change caused by FL becomes very subtle. MOCPD is only able to achieve a F2-score of 0.20 for 0.05 gph data as the lower leak rate is a challenging case to tackle.  
\begin{table}[]
\caption{Results on 0.1 gph FL data.}
\centering
\begin{tabular}{lllll}
\hline
Method            & Recall & Precision & F2 & Delay (day) \\ \hline
BOCD              & 0.3567 & 0.0955 & 0.2306 & 4.60  \\
M-Statistic       & 0.3322 & 0.2571 & 0.3138 & 5.11  \\
NEWMA             & 0.3158 & 0.0841 & 0.2036 & 3.57 \\
SEP               & 0.1696 & 0.0883 & 0.1432 & 6.68 \\
AE                & 0.2122 & 0.1666 & 0.2012 & 7.24 \\
OC                & 0.0930 & 0.2370 & 0.1059 & 2.95  \\
LIFEWATCH         & 0.3743 & 0.2977 & 0.3560 & 5.44 \\ \hline
\textbf{MOCPD-Mean}     & 0.4958 & 0.3274 & \textbf{0.4495} & 6.39  \\ 
\textbf{MOCPD-MMD}      & 0.4833 & 0.3345 & 0.4438 & 6.24\\ 
\textbf{MOCPD-VAE}      & 0.3760 & 0.3671 & 0.3742 & 5.25  \\ \hline
\end{tabular} 
\label{tab:01}
\end{table}

\begin{table}[]
\caption{Results on 0.05 gph FL data.}
\centering
\begin{tabular}{lllll}
\hline
Method            & Recall & Precision & F2 & Delay (day) \\ \hline
BOCD              & 0.1287 & 0.0345 & 0.0832 & 3.88 \\
M-Statistic       & 0.3626 & 0.0420 & 0.1436 & 4.07  \\
NEWMA             & 0.2645 & 0.0366 & 0.1179 & 4.20 \\
SEP               & 0.0429 & 0.0247 & 0.0374 & 6.67 \\
AE                & 0.1094 & 0.0781 & 0.1012 & 6.76 \\
OC                & 0.0204 & 0.0824 & 0.0240 & 3.58  \\
LIFEWATCH         & 0.2791 & 0.0802 & 0.1866 & 3.99 \\ \hline
\textbf{MOCPD-Mean}     & 0.2166 & 0.1714 & \textbf{0.2057} & 6.39 \\ 
\textbf{MOCPD-MMD}      & 0.2083 & 0.1712 & 0.1997 & 6.09\\ 
\textbf{MOCPD-VAE}      & 0.1542 & 0.1829 & 0.1591 & 5.38  \\ \hline
\end{tabular} 
\label{tab:005res}
\end{table}
\begin{figure*}[]
  \centering  
    \includegraphics[width=0.9\textwidth]{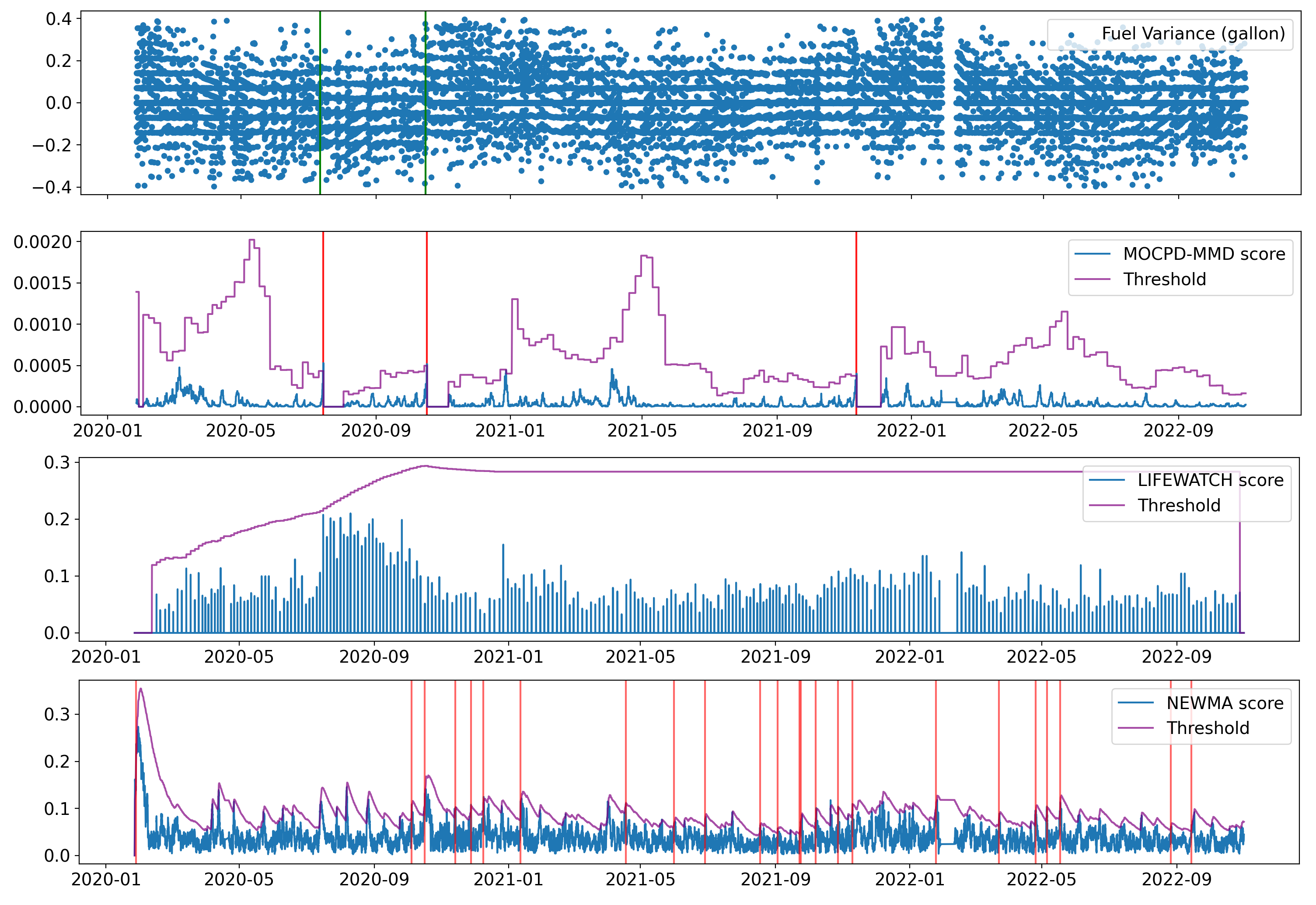}
  \caption{The dissimilarity scores and thresholds obtained by different algorithms on the 0.1gph FL sample (top plot). Green vertical lines indicate the ground truths where the leakage starts and is fixed. Red vertical lines denote the predictions made by the algorithms.}
  \label{fig:01scores}
\end{figure*}
Figure \ref{fig:01scores} shows an example of 0.1 gph fuel leakage data and the corresponding scores. Compared to the example in Figure \ref{fig:02scores}, the shift caused by FL in this tank sample is less obvious and humans can visually tell there are more gradual changes in the overall trend. Yet, MOCPD-MMD is able to capture the actual change points with only one false alarm. Although LIFEWATCH generates peaks near the ground truth, they do not exceed the threshold. On the contrary, NEWMA generates many false alarms and fails to capture the first change point.

\subsection{Real-World Fuel Leakage Data}
\begin{figure}
\centering
\subfloat[Tank A\label{fig:real1}]{\includegraphics[width=0.8\columnwidth]{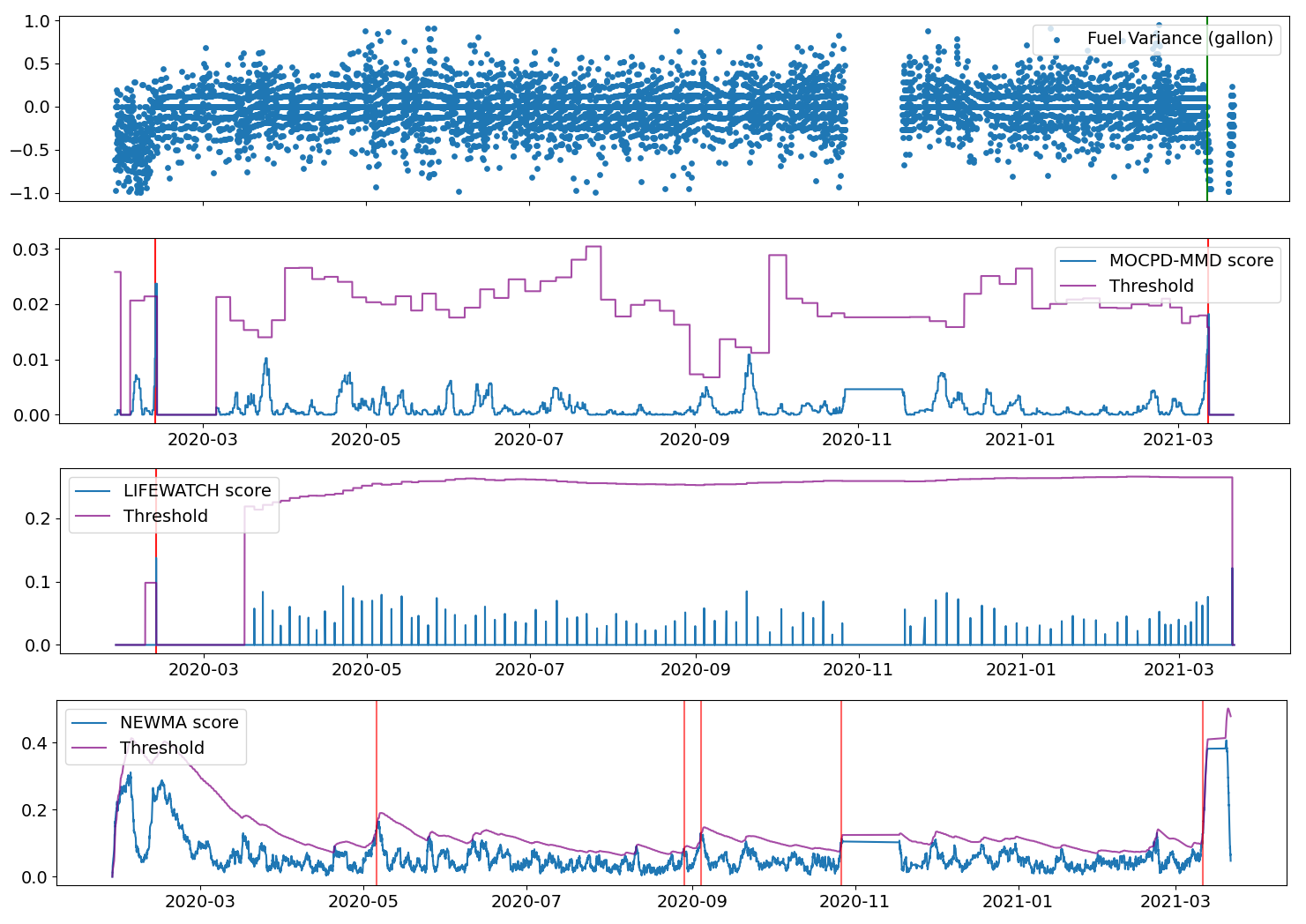}}\par 
\subfloat[Tank B\label{fig:real2}]{\includegraphics[width=0.8\columnwidth]{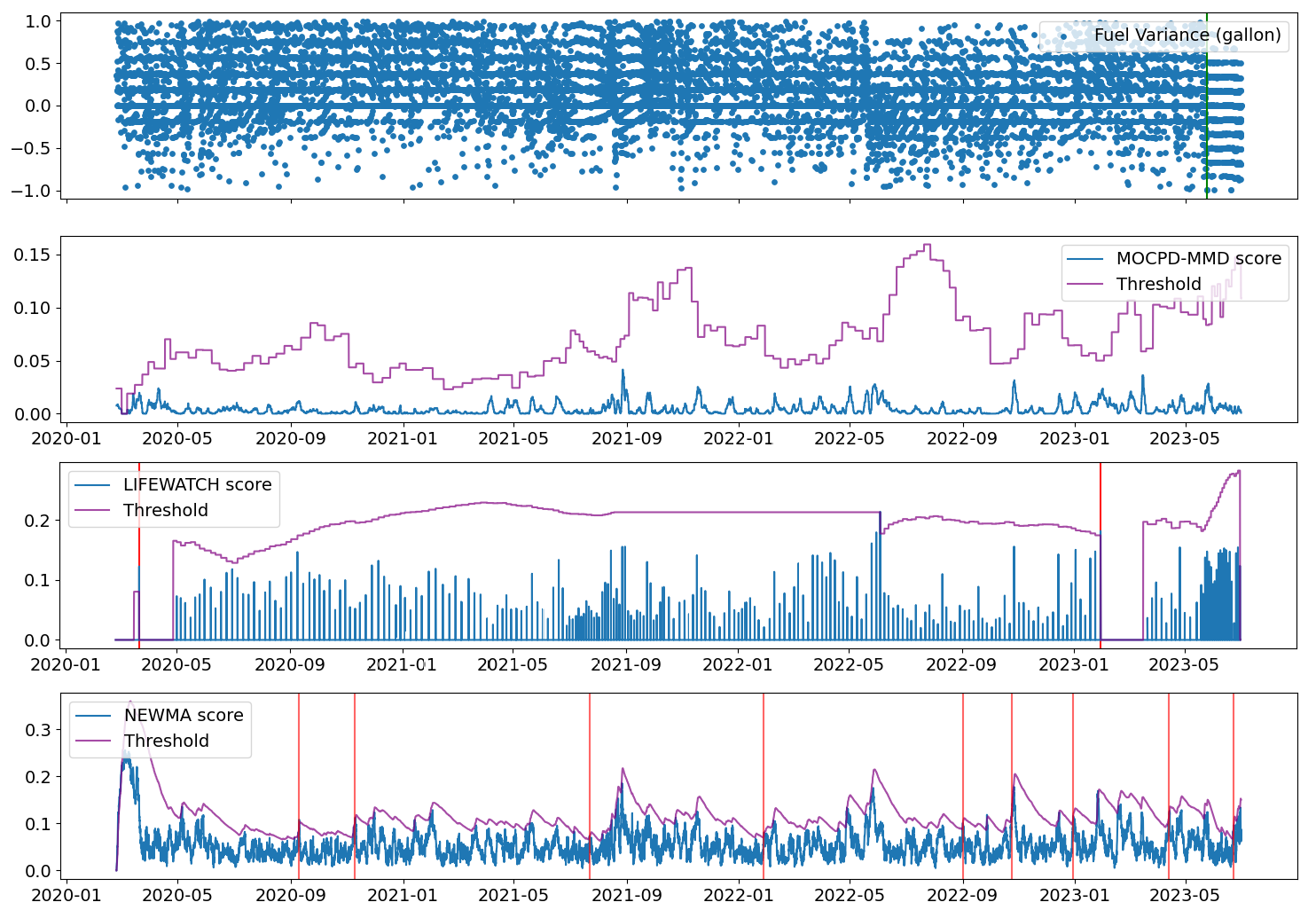}}
\caption{Real FL case study.}
\label{fig:real}
\end{figure}

In this section, we apply MOCPD-MMD as well as two baseline methods to the real-world FL data, to allow us to investigate how online CPD methods perform in real FL scenarios. We gather two real-world FL cases. However, their actual leak rates and the positions of the leakage are unknown. The input variance data and dissimilarity score generated by the methods are visualised in Figure \ref{fig:real}. MOCPD-MMD and NEWMA are able to pick up the leakage in Figure \ref{fig:real1} promptly. Similar to the observations in the simulated case study, NEWMA tends to produce a lot of false alarms, which is a major drawback of memory-free algorithms. Although false alarms are also common in practice, detection by a system becomes untrustworthy if it generates too many false alarms. For the case in Figure \ref{fig:real2}, all the methods fail to detect the leakage. One possible cause of the detection failure could be that the tank sample data are too noisy. These methods may struggle to distinguish actual leakage from other sudden local trend changes.

\subsection{Result Discussion on Benchmark Datasets}
In this section, we demonstrate the generality of MOCPD by evaluating its performance on two simulated and one real-world commonly used benchmark CPD datasets. They are simulated jumping mean and gaussian mixture \citep{liu2013change, chang2019kernel}, and HASC \citep{kawaguchi2011hasc}.

\textbf{Jumping Mean (JM)} a synthetic dataset generated using the autoregression model provided below:  
\begin{equation}
\begin{alignedat}{2} \label{eq:ar}
x_{i} = 0.6x_{i-1} - 0.5x_{i-2}  + \epsilon_{i} \\
\end{alignedat} 
\end{equation}
where $\epsilon_{i} \sim \mathcal{N}(0,1.5)$. A CP is created at every 500 time steps.The noise mean $\mu$ at time $i$ is set as: 
\begin{equation}
  \mu_{N}=\begin{cases}
    0, & \text{$N=1$}.\\
    \mu_{N-1} + \frac{N}{16}, & \text{$N=2,...,49$}.
  \end{cases}
\end{equation}
where $N$ is a natural number such that $500(N - 1) + 1\leq t \leq 500N$. The tolerance distance is set to 25 and window size is set to 25. Memory size is set to 10 for any method using a limited size of memory. 

\textbf{Gaussian Mixtures (GM)}. Time series data are sampled alternatively between two Gaussian mixtures: $0.5\mathcal{N}(-1, 0.5^{2}) + 0.5\mathcal{N}(1, 0.5^{2})$ and $0.8\mathcal{N}(-1, 1.0^{2}) + 0.2\mathcal{N}(1, 0.1^{2})$. CPs are generated following the same mechanism as JM and the parameter setting is the same as JM. 

\textbf{HASC\footnote[1]{http://hasc.jp/hc2011}} is a subset of the HASC Challenge 2011 dataset \citep{kawaguchi2011hasc}. It contains human activity data captured from portable three-axis accelerometers. Following the steps in \citet{liu2013change, li2015m}, we use a subset of the dataset, merge the activity segments and convert the data to one dimension by taking the l2-norm of the 3-dimensional data. The tolerance distance is set to 50, the window size is set to 50 and the memory limit is set to 20. 

\begin{table}[]
\caption{F1-scores on benchmark datasets.}
\centering
\begin{tabular}{llll}
\hline
Method            & JM & GM & HASC  \\ \hline
BOCD              & 0.0769 & 0.6286 & 0.2962 \\
M-Statistic       & 0.6118 & \textbf{0.6666} & \textbf{0.3512} \\
NEWMA             & 0.5769 & 0.6173 & 0.3428 \\
SEP               & 0.2667 & 0.6047 & 0.1538 \\
AE                & 0.2846 & 0.2099	& 0.1242 \\
OC                & 0.3636 & 0.1090 & 0.1667  \\
LIFEWATCH         & 0.6429 & 0.1967 & 0.2632 \\ \hline
\textbf{MOCPD-Mean}        & 0.4940 & 0.5223 & 0.3017 \\ 
\textbf{MOCPD-MMD}         & \textbf{0.6611} & 0.6422 & 0.3436 \\ 
\textbf{MOCPD-VAE}         & 0.1406 & 0.1682 & 0.1128 \\ \hline
\end{tabular} 
\label{tab:bench}
\end{table}
We use F1-score (Equation \ref{eq:fscore} with $\beta=1$) as the main criterion and results on three benchmark datasets are summarised in Table \ref{tab:bench}. It is worth noting that the benchmark datasets comprise a very limited number of sequence samples but each sequence contains many CPs (30$\sim$50). In contrast, our FL data consists of a greater number of sequences but much fewer CPs within each sequence. This also implies that the length of each state will be shorter compared to our problem setting. We can observe that MOCPD-MMD can still achieve competitive scores for benchmark datasets, as it is effective for various types of CPs. M-Statistic and NEWMA also show great performance, while OC has the worst performance. LIFEWATCH shows poor performance on GM, which is a dataset that alternates between two states. As LIFEWATCH stores all the detected distribution, it tends to perform badly when it makes a mistake in recognising the distribution.

\subsection{Further Discussion}
In this section, we summarise some key insights based on the experimental results presented in the previous sections, not only on FL detection but also on benchmark datasets. 

Overall, memory-based algorithms including our proposed method MOCPD and LIFEWATCH, show superior performance in the FL detection problem. This superiority stems from their abilities to store representative reference samples and employ an adaptive threshold mechanism. In the FL detection setting where states tend to be more long-lasting and transitions are subtler that could be masked by noises and local trends, comparing the latest window to the right selection of past reference samples is crucial to maintain detection accuracy. LIFEWATCH has worse performances compared to MOCPD as it does not update its memory once full, resulting in the omission of recent sample information. On the other hand, memory-free methods while being lightweight, are more sensitive to noise, making them less robust in our problem settings. Furthermore, given the great number of sequential instances with varying baselines, an adaptive threshold that is built based on each instance proves more effective than a standardised threshold applied uniformly across all tanks. Meanwhile, the performance of memory-based methods still relies on a sufficiently large memory size and an appropriate selection of dissimilarity measures.   

Through experiments with benchmark CPD datasets, we show that MOCPD is able to maintain competitive scores, highlighting its general applicability. However, it becomes less advantageous when transitions occur quickly with each state lasting for only a short period, leading to fewer samples. This is especially noticeable when MOCPD is used with VAE that its worsen performances could be caused by not having sufficient new samples for model updates. Adequate training data is essential for strong ML performance. Additionally, when transitions are clearly distinguishable from noise, all methods become capable of detecting changes, as noise has less impact. While MOCPD is specifically designed to be more robust to noise.  

\section{Conclusion}
In this paper, we introduce a novel memory-based online CPD framework MOCPD for real-time FL detection to address the detection delays common in existing FL detection methods. Through extensive experiments on real-world fuel variance data with induced leakages, we demonstrate the effectiveness of MOCPD in early FL detection within an online setting. MOCPD outperforms baseline online CPD methods by leveraging memory to retain representative historical experience. It periodically updates this memory with more recent data and adaptively adjusts the threshold to detect state changes. Parametric-based as well as ML-based dissimilarity measures have been evaluated in conjunction with MOCPD. MOCPD-Mean and MOCPD-MMD show superior performances on FL data, achieving much higher recall scores while maintaining reasonable precision scores with a much shorter detection delay compared to the general turnaround time of the methods in practice. Additionally, we conduct a sensitivity analysis on the window size, memory size limit and simulated leak rate to examine their impacts on models' performances. Through case studies on both simulated and real-world FL data, we show how MOCPD performs better than selected baselines; however, we also acknowledge the challenges of dealing with some real-world scenarios where trends gradually change over time which could mask the true changes. Finally, through experiments on benchmark CPD datasets, we demonstrate the efficacy and general applicability of the MOCPD framework as it can be combined with various types of dissimilarity measure methods. For future work, we plan to incorporate additional variables such as sales volume and formulate real-time FL detection as a multivariate problem. Smaller leak rates remain a challenging problem for early fuel leakage detection \citep{husain2022current}. Given that examined algorithms perform worse on smaller leak rate data and there are various noises in our data, we will further explore more effective methods to address these challenging issues. Lastly, we can investigate how the methods perform in a more complex scenario where leakage occurs at the middle or the top of the tank.

\section*{Acknowledgement}
This work is funded by the Australian Research Council (ARC) Linkage Grant (LP190100991). 

\bibliographystyle{unsrtnat}
\bibliography{RTFL} 

\end{document}